\pdfoutput=1

\documentclass[11pt]{article}

\usepackage[final]{acl}

\usepackage{times}
\usepackage{pifont}
\usepackage{booktabs}
\usepackage[table]{xcolor} 
\definecolor{headergray}{gray}{0.9} 

\usepackage{pifont}  
\usepackage{xcolor}  
\usepackage{colortbl} 

\definecolor{cgreen}{RGB}{46, 139, 87}  
\definecolor{cred}{RGB}{220, 50, 50}    

\newcommand{\yes}{\textcolor{cgreen}{\ding{51}}} 
\newcommand{\no}{\textcolor{cred}{\ding{55}}}    
\usepackage{multirow}
\usepackage{xcolor}
\usepackage{colortbl}
\usepackage{bm}
\usepackage{latexsym}

\usepackage[T1]{fontenc}

\usepackage[utf8]{inputenc}

\usepackage{microtype}

\usepackage{inconsolata}

\usepackage{graphicx}

\usepackage{amsmath}
\usepackage{amssymb}
\usepackage{multirow}
\usepackage{arydshln}
\usepackage{graphicx}
\usepackage{wrapfig}
\usepackage{colortbl}
\usepackage{booktabs}
\usepackage{makecell}

\usepackage{diagbox}
\usepackage{float}

\RequirePackage{xspace}
\makeatletter
\DeclareRobustCommand\onedot{\futurelet\@let@token\@onedot}
\def\@onedot{\ifx\@let@token.\else.\null\fi\xspace}

\title{Bridging the Pose-Semantic Gap: A Cascade Framework for  Text-Based Person Anomaly Search}

\author{
  \textbf{Zequn Xie}\textsuperscript{1,\textsuperscript{†}},
  \textbf{Guijin Luo}\textsuperscript{1,\textsuperscript{†}},
  \textbf{Chuxin Wang}\textsuperscript{1,\textsuperscript{†}},\\
  \textbf{Sihang Cai}\textsuperscript{1},
  \textbf{Tao Jin}\textsuperscript{1},
  \textbf{Zhou Zhao}\textsuperscript{1},
  \textbf{Yixuan Tang}\textsuperscript{2}\thanks{Corresponding author.}
  \\
  \normalsize \textsuperscript{1} Zhejiang University \\
  \normalsize \textsuperscript{2} National University of Singapore \\
  \normalsize Correspondence: \texttt{zqxie@zju.edu.cn}, \texttt{yixuan@comp.nus.edu.sg}
}
\begin{document}
\maketitle

\begin{abstract} Text-based person anomaly search retrieves specific behavioral events from surveillance archives using natural-language queries. Although recent pose-aware methods align geometric structures well, they face a fundamental \textit{Pose-Semantic Gap}: semantically different actions can share similar skeletal geometries. While Multimodal Large Language Models (MLLMs) can reduce this ambiguity, using them for large-scale retrieval is computationally prohibitive. We propose the \textit{Structure-Semantic Decoupled Cascade (SSDC)} framework, which decouples retrieval into two stages: (1) \textit{Structure-Aware Coarse Retrieval}, where a lightweight model quickly filters candidates by skeletal similarity; and (2) {Detective Squad Interaction}, a multi-agent semantic verification module. The squad consists of a \textit{Detective} for fast binary filtering, an \textit{Analyst} for evidence extraction, and a \textit{Writer} for semantic synthesis. Finally, we re-rank candidates by fusing the synthesized captions with structural priors. Experiments on the PAB benchmark show that SSDC achieves state-of-the-art performance by balancing efficiency and semantic reasoning.Our code is available at: \url{ https://github.com/GridNexus/SSDC}.
 \end{abstract}

\begin{figure}[htbp]
    \centering
    \includegraphics[width=0.5\textwidth]{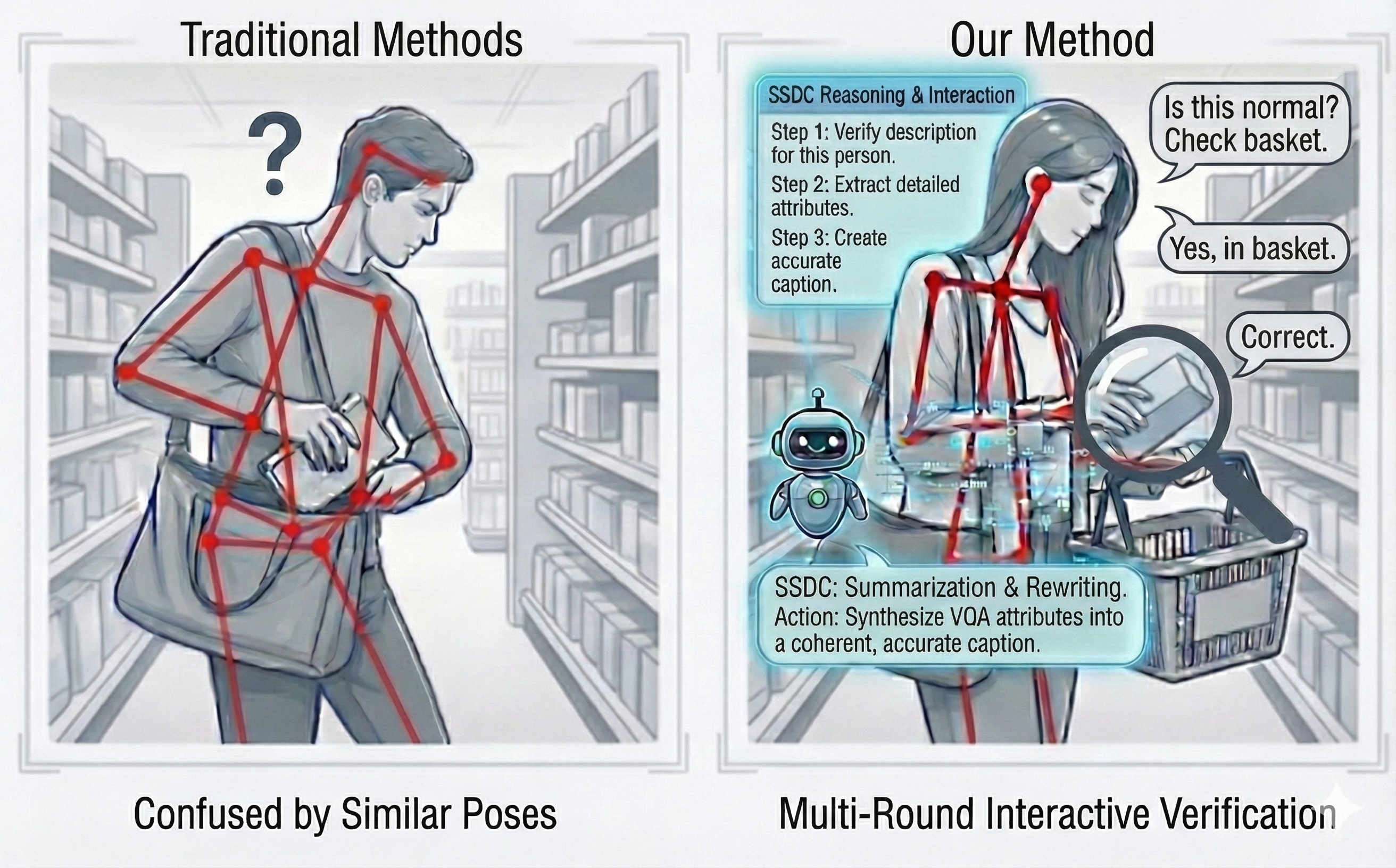}

\caption{Illustration of the Pose-Semantic Gap. Traditional pose-aware methods (left) fail to distinguish semantically distinct actions with similar skeletal geometries. Our SSDC framework (right) bridges this gap through Multi-Round Interactive Verification using a collaborative agent workflow (Verify 
$\rightarrow$
 Extract 
$\rightarrow$
 Rewrite) to analyze fine-grained visual details and generate semantically accurate captions. } 
    \label{fig:1}
\end{figure}

\begin{figure*}[htbp]
    \centering
    \includegraphics[width=1\textwidth]{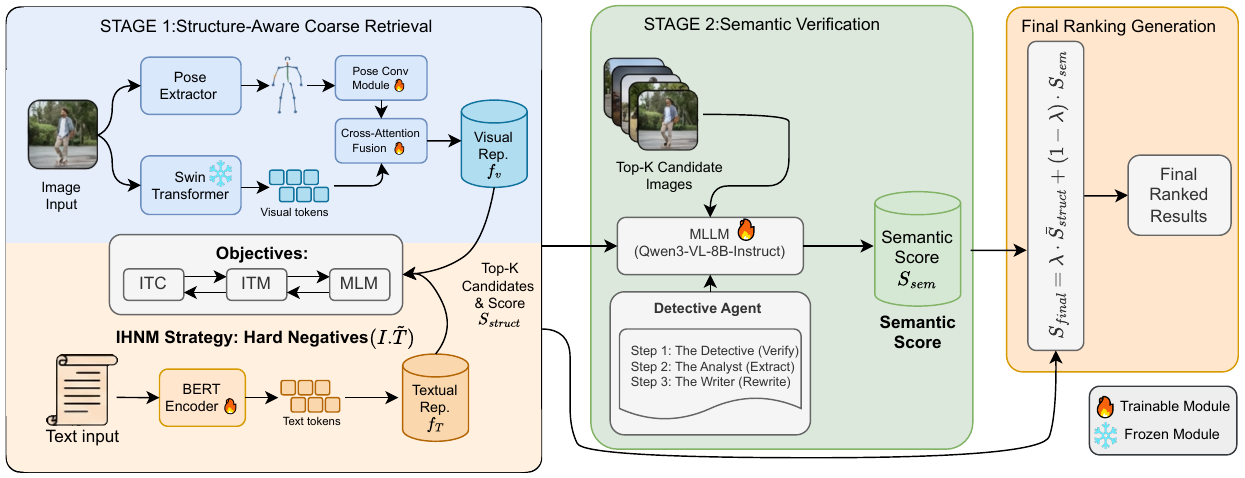}
    \caption{\textbf{Overall architecture of the SSDC framework .} The framework follows a coarse-to-fine pipeline : (1) Coarse Retrieval uses a lightweight model to filter the gallery based on structural similarity. (2) Semantic Verification introduces a specialized Detective Agent to scrutinize hard negatives. This agent employs Detective-style Prompting to resolve fine-grained ambiguities through multi-round reasoning and visual entailment. Finally, an Adaptive Fusion Mechanism integrates structural scores with the agent's semantic verdicts to produce the final ranking.}
    \label{fig:framework}
\end{figure*}

\section{Introduction}
\label{sec:intro}

Text-based person search~\cite{li2017person,zheng20242} has emerged as a critical technology in intelligent surveillance, allowing for the retrieval of specific individuals using natural language descriptions. While effective for routine identification, existing methods struggle to address the complexities of real-world security, where detecting anomalies is paramount. Consequently, a new task, \textit{text-based person anomaly search}, has been introduced. This task requires identifying pedestrians involved in both routine and anomalous activities. However, it faces a critical challenge: the {``Pose-Semantic Gap.''} Distinct behaviors, such as \textit{doing push-ups} versus \textit{falling down}, often exhibit nearly identical skeletal geometries. Traditional pose-aware models, which rely heavily on geometric alignment, fail to discern these subtle semantic distinctions, leading to high false-positive rates.

While Multimodal Large Language Models (MLLMs) possess the semantic reasoning capabilities necessary to resolve such ambiguities, deploying them directly on massive surveillance archives is computationally prohibitive. The sheer volume of video data demands high-throughput processing, yet MLLMs suffer from significant inference latency. This creates a dilemma: relying on lightweight models that sacrifice semantic precision for speed, or employing large models that offer accuracy but lack the efficiency required for real-time deployment.

To bridge this gap, we propose the {Structure-Semantic Decoupled Cascade (SSDC) Framework}. Our approach synergizes the efficiency of structural filtering with the reasoning power of large models. The retrieval process is decomposed into two synergistic stages: 
(1) A \textbf{Structure-Aware Coarse Retrieval} stage, where a lightweight Structure-Aware Coarse Retriever  acts as a high-speed filter to recall a candidate pool based on structural similarity; and 
(2) A \textbf{Detective Squad Interaction} stage, where we replace monolithic verification with a novel multi-agent collaboration. Unlike previous methods, we orchestrate three specialized agents: a \textbf{Detective} for rapid binary filtering to discard obvious negatives, an \textbf{Analyst} for structured evidence extraction via a checklist, and a \textbf{Writer} for synthesizing fragmented evidence into a comprehensive, high-quality caption. Finally, this refined semantic description is used for precision re-ranking.

To support this task and validate our framework, we utilize the PAB benchmark~\cite{yang2025beyondcmp}. This large-scale dataset comprises over 1 million image-text pairs, encompassing 1,600 anomaly types and 1,000 normal action types. In summary, our primary contributions are:

\begin{itemize}
    \item We propose the SSDC Framework, a coarse-to-fine architecture that effectively bridges the Pose-Semantic Gap by decoupling structural filtering from semantic verification, balancing retrieval efficiency with reasoning depth.

\item We design the \textbf{Detective Squad}, a multi-agent verification mechanism with three roles: {Detective} (Filter), {Analyst} (Extractor), and {Writer} (Integrator). Notably, this framework serves as a \textbf{plug-and-play} module that seamlessly adapts to various retrieval backbones. 

    \item Extensive experiments on the PAB benchmark show that our method significantly outperforms SOTA baselines, validating our agent-based collaborative strategy.
\end{itemize}

\section{Related Work}

\subsection{Text-Based Person Anomaly Search.}
This task sits at the intersection of person retrieval and anomaly detection, requiring the localization of specific pedestrians based on fine-grained behavioral descriptions. {In the domain of Text-Based Person Retrieval}, methods have evolved from early global alignment~\cite{zheng2020dual, ding2021semantically} to precise local matching via attention mechanisms~\cite{wang2022look, shu2023see}. Recently, Vision-Language Pre-training (VLP) models like IRRA~\cite{jiang2023cross} and RaSa~\cite{bai2023rasa} have achieved state-of-the-art results by leveraging CLIP~\cite{radford2021learning} for robust representation. Further, recent works have explored uncertainty modeling to handle noisy text-image correspondences~\cite{xie2025dynamic}, and incorporated highly controllable prompt learning architectures to refine cross-modal representation matching~\cite{yang2025multimodal}. \textit{However}, these approaches primarily target {static appearance attributes} and overlook complex behavioral semantics, leading to a ``Pose-Semantic Gap'' where geometrically similar actions are misidentified. While some recent temporal detection frameworks attempt to enhance spatiotemporal generalization across diverse scenes~\cite{Feng_Cai_Xie_Wu_Jin_2026}, their application to fine-grained, text-queried person search remains under-explored. {Conversely, in the realm of Person Anomaly Detection}, traditional Video Anomaly Detection (VAD) focuses on identifying statistical deviations in temporal sequences, often employing one-class classification~\cite{zaheer2022generative, flaborea2023multimodal} or weakly supervised ranking~\cite{sultani2018real}. While effective for general monitoring, these methods lack the {semantic flexibility} to handle open-vocabulary natural language queries. Although recent works like UCA~\cite{yuan2024towards} attempt to incorporate text, they operate at a coarse \textit{video level} rather than the \textit{instance level}. Consequently, neither domain alone solves the challenge of high-precision, instance-level anomaly retrieval.

\subsection{Synergizing Structure and Semantics.}
Integrating structural priors with semantic reasoning is a growing direction for disambiguating complex human behaviors. Pose-guided methods~\cite{jing2020pose, zhu2021dssl} utilize keypoints to improve feature alignment. \textit{However}, pose information alone remains ambiguous; distinct actions like \textit{falling} and \textit{push-ups} share identical skeletal geometries, leading to false positives when semantic context is absent. On the other hand, MLLMs demonstrate superior capabilities in semantic reasoning and visual entailment. While some studies utilize MLLMs for synthetic data generation~\cite{yang2023towards, xie2025chat} or auxiliary supervision~\cite{tan2024harnessing}, directly deploying them for large-scale retrieval is computationally prohibitive due to high inference latency. Furthermore, ensuring accurate spatiotemporal reasoning with large vision-language models requires careful mitigation of structural hallucinations, a challenge recently addressed through progressive training strategies in complex embodied scenarios~\cite{yang2026progressive}. Recent advancements have sought to alleviate these bottlenecks and enhance the reliability of large vision-language models in cross-modal retrieval through uncertainty-aware inference~\cite{gong2026towards}. Despite these efforts, existing pipelines often employ simple re-ranking strategies without explicitly modeling the structure-semantic discrepancy. In contrast, our Structure-Semantic Decoupled Cascade (SSDC) Framework uniquely synergizes these paradigms. We utilize a lightweight pose-aware model for efficient {Coarse Retrieval} and employ a specialized {Detective Agent} for {Semantic Verification}. This decoupled design effectively bridges the Pose-Semantic Gap, achieving a superior balance between retrieval accuracy and efficiency~\cite{xie2026conquer, xie2026hvd, xie2026delving}.

\begin{figure*}[t]
  \centering
   \includegraphics[width=\linewidth]{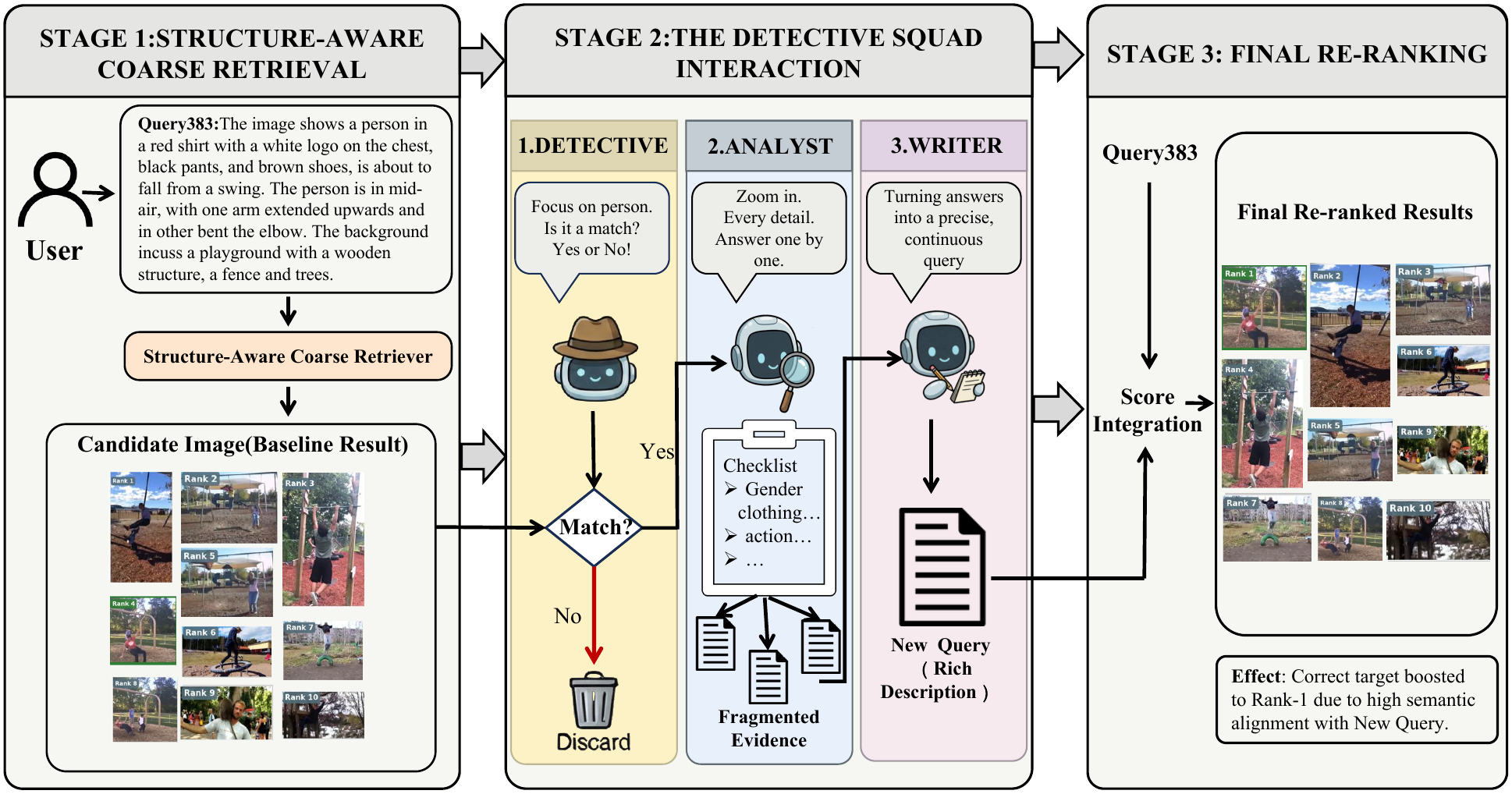}
  \caption{Overview of the proposed Detective Squad framework for person re-identification. The pipeline operates in a coarse-to-fine manner across three stages. Stage 1 utilizes a lightweight model for initial Structure-Aware Coarse Retrieval to obtain baseline candidates. Stage 2 introduces the novel multi-agent Detective Squad (Detective, Analyst, and Writer) to perform semantic verification, filter distractors, extract fine-grained details, and generate precise new captions. Stage 3 executes Final Re-ranking by fusing scores based on the generated captions, significantly boosting the rank of the target image. }
  \label{fig:result}
\end{figure*}

\section{Method}
\label{sec:method}

As illustrated in Figure~\ref{fig:framework}, we propose the {Structure-Semantic Decoupled Cascade (SSDC) Framework}, a coarse-to-fine approach that bridges the ``Pose-Semantic Gap'' in text-based person anomaly search. Traditional single-stage models often struggle to distinguish semantically distinct behaviors with similar skeletal geometries. SSDC therefore decouples retrieval into two stages. First, a lightweight {Structure-Aware Coarse Retriever} filters the gallery by structural similarity to retrieve a candidate pool with high recall and low latency. Second, {Detective Squad Interaction} performs semantic verification. Unlike monolithic approaches, we use a collaborative multi-agent workflow: a {Detective} discards clear negatives, an {Analyst} extracts fine-grained visual evidence, and a {Writer} synthesizes it into a precise \textit{new Query}. We then re-rank candidates using this refined description to improve semantic accuracy.

\subsection{Structure-Aware Coarse Retriever}
\label{sec:stage1}
The primary goal of Stage I is to recall a high-quality candidate pool $\mathcal{C}$ from the gallery $\mathcal{G}$ ($|\mathcal{C}| \ll |\mathcal{G}|$). We adopt the Cross-Modal Pose-aware framework~\cite{yang2025beyondcmp} as the backbone.

\paragraph{Pose-aware Representation Learning.}
Anomalous behaviors often correspond to distinctive body configurations. We extract a pose map $P$ from the input image $I$ and encode it with a lightweight Pose Conv Module, while a Vision Transformer processes $I$ to produce patch-level embeddings $f_I$. We then fuse pose and image features via \textit{Pose-aware Cross-Attention}: normalized pose features $f_P$ act as queries that attend to image features $f_I$, yielding a structure-enhanced representation $f_{CA}$:
\begin{equation}
    f_{CA} = \text{Softmax}\left(\frac{(W_q f_P)(W_k f_I)^T}{\sqrt{d}}\right) (W_v f_I).
\end{equation}
Finally, we obtain the visual embedding through a residual connection: $f_V = f_I + f_{CA}$.

\paragraph{Training Objectives.}
The coarse retriever is optimized using a hybrid objective: $\mathcal{L}_{stage1} = \mathcal{L}_{itc} + \mathcal{L}_{itm} + \mathcal{L}_{mlm}$. We employ {Identity-based Hard Negative Mining } to construct triplets where negative samples share the same identity but perform different actions, compelling the model to decouple action-specific semantics from identity appearance.

\subsection{The Detective Squad Interaction}
\label{sec:stage2}
While Stage I efficiently filters structurally irrelevant samples, it lacks the reasoning depth to resolve fine-grained ambiguities. To bridge this gap, we introduce the \textbf{Detective Squad}, a multi-agent framework designed to conduct hierarchical interactive reasoning on the top-$K$ candidates.

\subsubsection{Constructing the Detective Squad}
General-purpose MLLMs often lack sensitivity to specific anomaly definitions. We bridge this domain gap through a Hard-Negative Aware Supervised Fine-Tuning (SFT) strategy.

\paragraph{Mining Hard Negatives.}
A competent detective must distinguish the true target from mimics. We simulate this by mining \textit{structural hard negatives} from the frozen Stage I retriever. For a query $T$, we identify images $I^{-}$ that are ranked high due to structural similarity but are semantically incorrect. These tricky suspects force the agents to look beyond skeletal geometry. We subsequently aggregate these samples to construct a specialized dataset $\mathcal{D}_{sft}$, which serves as the foundation for training our agents to distinguish anomalies from routine actions.

\paragraph{Role-Specific Instruction Tuning.}
Instead of a simple binary task, we formulate verification as a \textit{Collaborative Visual Reasoning} problem. We construct instruction templates that correspond to three specialized roles: (1) A \textbf{Detective} tailored for binary filtering (Match or Discard); (2) An \textbf{Analyst} trained to answer a fine-grained checklist (e.g., ``Gender, Clothing, Action''); and (3) A \textbf{Writer} optimized to synthesize these details into coherent captions. We employ Low-Rank Adaptation (LoRA) to efficiently optimize the backbone, ensuring each agent adheres to its specific persona.

\subsubsection{Detective-style Inference Chain}
Standard captioning often yields hallucinations. To mitigate this, we propose a Detective-style Prompting mechanism that guides the squad's sequential reasoning process.

\paragraph{Hypothesis Testing via Role-Playing.}
We inject system prompts that assign specific personas to the agents, triggering a chain-of-thought verification.
First, the \textbf{Detective} acts as a filter, asking ``Is it a match? Yes or No!'' to discard obvious distractors.
Second, for surviving candidates, the \textbf{Analyst} performs a ``physical examination,'' checking attributes against a 15-point checklist to extract fragmented evidence.
Finally, the \textbf{Writer} acts as an integrator, synthesizing these verified details into a precise, continuous new caption $T_{new}$.

\paragraph{Generative Semantic Scoring.} 
Unlike opaque probability logits, our method produces an interpretable semantic anchor. We verify the consistency between the original query $T$ and the Writer's synthesized caption $T_{new}$ by computing their cosine similarity in the feature space:
\begin{equation}
    S_{sem}(I_k, T) = \frac{E_{txt}(T) \cdot E_{txt}(T_{new})}{\|E_{txt}(T)\| \|E_{txt}(T_{new})\|}
\end{equation}
where $E_{txt}$ denotes the frozen text encoder (e.g., from the CLIP or BERT backbone) used to extract semantic embeddings. This explicit text-to-text matching score $S_{sem}$ quantifies the semantic entailment, serving as the input for the subsequent fusion stage.

\paragraph{Generative Semantic Re-ranking.}
Instead of opaque probability scores, our method outputs explicit semantics. The \textit{new Query} generated by the Writer serves as a semantic anchor. We compare it with the original query to compute a high-fidelity semantic score, bridging the pose-semantic gap through interpretable text generation rather than black-box logits.

\subsubsection{Efficiency-Aware Dynamic Inference}
To apply the Detective Squad at scale, efficiency is critical. Inspired by cascade designs, we propose a Threshold-Gated Interaction mechanism.

\paragraph{Dynamic Activation Strategy.}
We observe that retrieval quality correlates with Stage I structural similarity. Candidates with low structural scores are likely irrelevant, making detailed analysis unnecessary.
Therefore, the {Detective} agent serves as an efficient gatekeeper. It is activated only for top-ranked candidates. If the Detective outputs ``No'', the sample is immediately discarded. This strategy effectively filters out distinct negatives, allowing the computationally more expensive {Analyst} and {Writer} to focus strictly on ambiguous candidates .

\subsection{Adaptive Fusion Mechanism}
The final ranking is determined by fusing structural priors with semantic reasoning. 
The semantic score $S_{sem}$ is calculated by measuring the textual similarity between the user's original query $T$ and the \textit{New Caption} $T_{new}$ synthesized by the Writer agent from the candidate image $I_k$:
\begin{equation}
    S_{sem}(I_k, T) = \frac{E_{txt}(T) \cdot E_{txt}(T_{new})}{\|E_{txt}(T)\| \|E_{txt}(T_{new})\|}
\end{equation}
where $E_{txt}$ denotes the frozen text encoder (e.g., BERT or CLIP text tower) used to extract semantic embeddings.

To balance efficiency and accuracy, we propose a threshold-gated fusion strategy. The final similarity score $S_{final}$ is computed as:
\begin{equation}
    S_{final} = 
    \begin{cases} 
    \lambda \tilde{S}_{str} + (1-\lambda) S_{sem}, & \tilde{S}_{str} > \xi \\
    \tilde{S}_{str}, & \text{else}
    \end{cases}
    \label{eq:fusion}
\end{equation}
where $\tilde{S}_{str}$ is the min-max normalized structural similarity score from Stage I, and $\xi$ is the confidence threshold. This ensures that the computationally intensive semantic verification is reserved strictly for high-potential candidates.
\begin{table*}[t]
  \centering
  \footnotesize
  \setlength{\tabcolsep}{2.5pt}
  \resizebox{0.99\linewidth}{!}{
  \begin{tabular}{@{}l|cc|cc|cc|cc|cc|cc|cc|cc|cc|cc@{}}
    \toprule
    \multirow{2}{*}{\large \textbf{Method}} & \multicolumn{2}{c|}{\textbf{Normal}} & \multicolumn{2}{c|}{\textbf{Wind}} & \multicolumn{2}{c|}{\textbf{Rain}} & \multicolumn{2}{c|}{\textbf{Snow}} & \multicolumn{2}{c|}{\textbf{Rain+Snow}} & \multicolumn{2}{c|}{\textbf{Dark}} & \multicolumn{2}{c|}{\textbf{Dark+Wind}} & \multicolumn{2}{c|}{\textbf{Dark+Rain}} & \multicolumn{2}{c|}{\textbf{Dark+Snow}} & \multicolumn{2}{c}{\textbf{Mean}$\bm{\uparrow}$} \\
    \cmidrule(lr){2-3} \cmidrule(lr){4-5} \cmidrule(lr){6-7} \cmidrule(lr){8-9} \cmidrule(lr){10-11} \cmidrule(lr){12-13} \cmidrule(lr){14-15} \cmidrule(lr){16-17} \cmidrule(lr){18-19} \cmidrule(lr){20-21}
    & \small R@1 & \small mAP & \small R@1 & \small mAP & \small R@1 & \small mAP & \small R@1 & \small mAP & \small R@1 & \small mAP & \small R@1 & \small mAP & \small R@1 & \small mAP & \small R@1 & \small mAP & \small R@1 & \small mAP & \small R@1 & \small mAP \\
    \midrule
    Baseline & 84.28  &91.28& 79.02 & 88.10 & 54.40 & 67.40 & 59.10 & 72.34 & 49.95 & 62.09 & 79.58 & 88.24 & 75.53 & 85.47 & 34.93 & 45.98 & 50.25 & 63.32
    & 63.00
 & 73.80 \\



    {IRRA} ~\cite{IRRA} 
    & {78.67} & {87.74} 
    
    & {75.08} &{85.40}
    
    &{56.62} &{ 70.37}

    &{ 58.34} &{72.06 }

    &{52.68 } &{  66.04}

    &{75.78 } &{  85.73}

    &{71.64  } &{82.95  } 

    & 38.32 & 50.03

    &{ 51.82 } &{ 64.66  }

    &{62.11 } &{73.89 } \\

    {CMP } ~\cite{yang2025beyondcmp} & {84.93} & {91.66} & {81.24} &{89.34} &{60.06} &{72.53} &{63.40} &{75.74} &{54.85} &{67.31} &{80.89} &{89.00} &{77.20} &{86.55} & 39.03 & 50.58 &{53.49} &{66.12} &{66.12} &{76.54} \\


    {RDE } ~\cite{qin2024noisy} 
    & {76.74} & {86.12} 
    
    & {  72.24} &{83.13}
    
    &49.90  &{  64.40 }

    &{ 53.03} &{66.69}

    &{ 46.21  } &{ 59.71}

    &{71.79} &{82.74 }

    &{67.54   } &{79.24  } 

      & 31.55 & 41.54 

    &{ 46.21 } &{ 58.47  }
    &{54.81 } &{66.99} \\ 

    {RDE + Detective Squad Interaction}
    
    &84.88   & 91.08
    
   & {80.59} & {88.09} 
    
    &{ 65.32 } &{  74.20}

    &{65.93 } &{ 74.81 }

    &{61.17 } &{  69.57}

    &{75.78 } &{  85.73}

    &{77.50  } &{85.39   } 

    & 41.61 & 47.48

    &{56.88 } &{ 64.88  }
    &{65.60 } &{73.77 } \\

    
    {IRRA + Detective Squad Interaction} 

    & {84.93} & {91.66} 
    
    & {81.65 } &{88.88}
    
    &{ 66.68  } &{  76.06  }
    
    &{ 67.85} &{77.44  }

    &\textbf{ 64.00 } &\textbf{72.76}

    &{ 81.50 } &{88.83 }

    &{79.12} &{ 87.10}

    &\textbf{49.04}   &   \textbf{56.48} 

    &{  61.07} &{  69.54 }

    &{70.65 } &{78.75 } \\

        \rowcolor{gray!10}
    \textbf{SSDC } 
    & \textbf{ 87.01 } & \textbf{ 92.74} 
    
    & \textbf{84.78} & \textbf{ 91.22 }    

    & \textbf{67.90 } & \textbf{76.68}     

    & \textbf{69.36} & \textbf{78.68}    

    & {62.29 } & {71.33}    

    & \textbf{82.86 } & \textbf{90.10}    

    & \textbf{80.59 } & \textbf{87.88}     

    &  {45.10}         &  {51.47}    

    & \textbf{ 61.78 } & \textbf{70.03}    

& \textbf{71.30} & \textbf{78.90} \\  
    
    \bottomrule
  \end{tabular}
  }
  \caption{Robust text-based person anomaly retrieval results on PAB under multi-weather setting. }
  \label{tab:multi}
  \vspace{-0.15in}
\end{table*}

\section{Experiments}
\label{sec:experiments}

In this section, we conduct extensive experiments on public benchmarks to evaluate the effectiveness, superiority, and generalization of SSDC.

\subsection{Experimental Setup}

 \noindent \textbf{Datasets.} We conduct experiments on the {Pedestrian Anomaly Behavior (PAB)} benchmark~\cite{yang2024beyond}, a large-scale dataset comprising over 1 million image-text pairs covering 1,600 anomaly types and 1,000 normal action types. To evaluate robustness against environmental variations, we employ a {Multi-weather Setting} that simulates 10 distinct weather conditions, mimicking round-the-clock smart city scenarios. Finally, to assess out-of-distribution generalization, we evaluate on the {UCC} test set derived from the real-world UCF-Crime dataset~\cite{sultani2018real}, which contains 5,320 unseen image-text pairs.

\noindent \textbf{Evaluation Metrics.} 
Following standard retrieval protocols~\cite{li2017person}, we report {Recall@K (R@1, R@5, R@10)} and mAP. A successful retrieval requires the top-ranked result to perfectly align with the query in terms of appearance, action intent, and background context.

 \noindent \textbf{Implementation Details.} All models are implemented in PyTorch. 
 
 In Stage I, we train the Structure-Aware Coarse Retriever  on a single RTX 3090 GPU for 30 epochs with a batch size of 22 using AdamW; the learning rate decays linearly from $1\times10^{-4}$ to $1\times10^{-5}$. 

\noindent In Stage II (\textbf{Detective Squad Interaction}), we use a cloud instance with a single NVIDIA A6000 GPU and BF16 precision. We adopt Qwen3-VL-8B-Instruct \cite{bai2025qwen3vltechnicalreport}as the unified backbone and fine-tune it with LoRA on our {multi-role instruction dataset} (9,000 samples) for 2 epochs. The dataset includes binary verification (Detective) and attribute checking (Analyst)  instructions. We apply LoRA to all linear modules with rank 8, alpha 16, and dropout 0.05. We optimize with a cosine scheduler (peak learning rate  $5.0 \times 10^{-5}$ , warmup ratio 0.1) and an effective batch size of 16. During inference, we use \textit{Efficiency-Aware Dynamic Inference}: the Detective gates subsequent reasoning, and we fuse the final scores with 
$\lambda=0.4$.

\subsection{Comparison with State-of-the-Art.} 
Table~\ref{tab:result} summarizes the quantitative performance of our SSDC framework alongside various SOTA methods, including general VLP models (e.g., CLIP~\cite{radford2021learning}, X-VLM~\cite{xvlm}) and specialized person retrieval methods (e.g., IRRA~\cite{jiang2023cross}, RaSa~\cite{bai2023rasa}). 

Under the zero-shot setting, general models exhibit limited adaptability; for instance, CLIP only reaches 47.57\% R@1. This confirms that standard semantic alignment struggles to bridge the specialized ``Pose-Semantic Gap'' in anomaly search. When fine-tuned on the 0.1M PAB dataset, our Stage I pose-aware retriever ({CMP}~\cite{yang2025beyondcmp}) already achieves {83.06\%} R@1 and {90.41\%} mAP. It outperforms established baselines such as X-VLM (81.95\% R@1) and RaSa (80.79\% R@1), validating the effectiveness of incorporating structural pose priors. 

With the full 1M training set, our complete {SSDC} framework sets a new state-of-the-art across all primary metrics, reaching \textbf{87.01\%} R@1 and \textbf{92.74\%} mAP. This significantly surpasses both the strong IRRA baseline (78.67\% R@1) and the standalone CMP model (84.93\% R@1). Notably, integrating the \textbf{Detective Squad Interaction} into a simpler backbone (IRRA + Detective Squad) enhances the R@1 to 83.92\%. This demonstrates that our collaborative verification stage effectively resolves hard ambiguity cases through fine-grained evidence extraction, leading to the best overall performance regardless of the base retriever.

\begin{table*}[t]
  \centering
  \footnotesize
  \setlength{\tabcolsep}{0pt} 
  \begin{tabular*}{\textwidth}{@{\extracolsep{\fill}}llcccccc@{}}
    \toprule
    \textbf{Method} & \textbf{Ref.} & \textbf{Image Enc.} & \textbf{Text Enc.} & \textbf{R@1} & \textbf{R@5} & \textbf{R@10} & \textbf{mAP} \\
    \midrule
    \multicolumn{8}{c}{\textit{Zero-shot Performance}} \\
    \midrule
    MRA~\cite{yang2025minimizing}   & CVPR'25   & Swin-B & BERT-Base & 9.91  & 23.66 & 31.45 & 17.15 \\
    RaSa~\cite{bai2023rasa}         & IJCAI'23  & ViT-B/16 & BERT-Base & 21.74 & 27.30 & 27.96 & 24.35 \\ 
    WoRA~\cite{sun2025data}         & WWW '25   & Swin-B & BERT-Base & 22.25 & 45.91 & 53.54 & 33.39 \\
    APTM~\cite{yang2023towards}     & MM'23     & Swin-B & BERT-Base & 22.90 & 45.80 & 52.38 & 33.56 \\
    CAMeL~\cite{yu2025camel}        & CVPR'25   & SG-Former & BERT-Base & 24.47 & 50.00 & 58.75 & 36.75 \\
    IRRA~\cite{jiang2023cross}      & CVPR'23   & ViT-B/16 & Transformer & 30.59 & 59.61 & 68.91 & 44.41 \\
    CLIP~\cite{radford2021learning} & ICML'21   & ViT-B/16 & Transformer & 47.57 & 81.55 & 89.03 & 62.73 \\  
    X-VLM~\cite{xvlm}               & ICML'22   & Swin-B   & BERT-Base & 71.94 & 97.78 & 98.99 & 83.96 \\

    RDE~\cite{qin2024noisy}      & CVPR'24   & ViT-B/16 & Transformer &  41.30 & 69.01 & 77.35 & 54.36 \\
    
    {IRRA + Detective Squad Interaction} & - & {ViT-B/16} & {Transformer} &   60.57 & 75.08  &  81.45  &68.00   \\
        {RDE + Detective Squad Interaction} & - & {ViT-B/16} & {Transformer} &   63.55 & 78.56   &   81.50 &71.00   \\

    \midrule

    \multicolumn{8}{c}{\textit{Fine-tuned with 0.1M PAB}} \\
    \midrule
    MRA~\cite{yang2025minimizing}   & CVPR'25   & Swin-B & BERT-Base & 70.53 & 94.69 & 97.47 & 81.59 \\
    APTM~\cite{yang2023towards}     & MM'23     & Swin-B & BERT-Base & 72.14 & 95.30 & 97.17 & 82.78 \\
    CAMeL~\cite{yu2025camel}        & CVPR'25   & SG-Former & BERT-Base & 74.30 & 96.79 & 98.84 & 84.20 \\
    WoRA~\cite{sun2025data}         & WWW '25   & Swin-B & BERT-Base & 74.47 & 96.82 & 98.48 & 84.60 \\
    IRRA~\cite{jiang2023cross}      & CVPR'23   & ViT-B/16 & Transformer & 76.39 & 97.62 & 99.14 & 86.33 \\
    CLIP~\cite{radford2021learning} & ICML'21   & ViT-B/16 & Transformer & 77.60 & {98.84} & \textbf{99.75} & 87.35 \\
    RaSa~\cite{bai2023rasa}         & IJCAI'23  & ViT-B/16 & BERT-Base & 80.79 & \underline{98.89} &\underline{ 99.65} & 89.20 \\
    X-VLM~\cite{xvlm}               & ICML'22   & Swin-B   & BERT-Base & 81.95 & { 98.84} & 99.19 & 89.86 \\
    CMP           ~\cite{yang2025beyondcmp}                    &ICCV'25          & ViT-B/16 & BERT-Base & \underline{83.06} & \underline{98.89} & 99.49 & \underline{90.41} \\
    \midrule
    \multicolumn{8}{c}{\textit{Fine-tuned with 1M PAB}} \\
    \midrule

    IRRA~\cite{jiang2023cross}      & CVPR'23   & ViT-B/16 & Transformer & 78.67 & 97.98 &98.94 & 87.74 \\
    
    CMP  ~\cite{yang2025beyondcmp} & ICCV'25 & ViT-B/16 & BERT-Base & 84.93 & \textbf{99.09} & \textbf{99.75} & 91.66 \\

    RDE~\cite{qin2024noisy}      & CVPR'24   & ViT-B/16 & Transformer &76.74 &  96.97 &98.38  & 86.12 \\
    RDE + Detective Squad Interaction & - & {ViT-B/16} & {Transformer} &  84.88 &98.13  &  98.69   & 91.08 \\

    {IRRA + Detective Squad Interaction} & - & {ViT-B/16} & {Transformer} &  83.92 &98.33 &  99.09  &90.60 \\
    
    \textbf{SSDC} & - & \textbf{ViT-B/16} & \textbf{BERT-Base} & \textbf{ 87.21 } & \textbf{99.09} & \textbf{99.75} & \textbf{92.87} \\

    \bottomrule

  \end{tabular*}
\caption{Quantitative results of our proposed method and compared methods on the PAB benchmark. \textbf{Bold} indicates the best result, and \underline{underlined} indicates the second best.}
  \label{tab:result}
  \vspace{-0.1in}
\end{table*}
\begin{table}[t]
  \centering
  
  \resizebox{\linewidth}{!}{%
    \begin{tabular}{@{}l|cccc@{}}
      \toprule
      Method & R@1 & R@5 & R@10 & mAP \\
      \midrule
      APTM~\cite{yang2023towards} & 27.86 & 40.41 & 46.77 & 22.61 \\
      CLIP~\cite{radford2021learning} & 51.60 & 68.31 & 76.43 & 43.05 \\
      X-VLM~\cite{xvlm} & 52.33 & 66.73 & 72.54 & 40.87 \\
      RaSa~\cite{bai2023rasa} &{54.12} & 70.32 & 75.96 & 39.71 \\
      IRRA~\cite{jiang2023cross} & 40.28 & 57.24 & 65.98 & 33.53 \\
      RDE~\cite{qin2024noisy} &  32.69 &48.55 &56.64 &  27.42 \\

      CMP~\cite{yang2025beyondcmp} &  \underline{55.23} &  \underline{71.67} &  \underline{77.99} &  \underline{44.35} \\
      \midrule
       RDE + Detective Squad Interaction & {41.58} & 48.52 & 56.39 &  28.20  \\
     IRRA + Detective Squad Interaction & {49.00} & {58.84} & {67.08} & {34.69} \\
      SSDC  & \textbf{59.45} & \textbf{72.68} & \textbf{78.30} & \textbf{45.25} \\

      \bottomrule
    \end{tabular}%
  }
  
  \vspace{-.1in}
  \caption{Comparisons with existing methods in OOD setting. The unseen test set UCC is extracted from the UCF-Crime~\cite{sultani2018real} dataset.}
  \vspace{-.1in}
  \label{tab:ood}
\end{table}

\noindent \textbf{Robustness to Environmental Variations.} 
Real-world surveillance often suffers from visual degradation. We evaluate model robustness across 10 distinct weather conditions (e.g., rain, snow, dark), as shown in Table~\ref{tab:multi}. SSDC demonstrates consistent performance gains across all scenarios, proving that the Detective Squad's hierarchical reasoning provides resilience even when visual structural cues are compromised.

\noindent \textbf{Out-of-Distribution Generalization.} To evaluate generalization, we test on the unseen UCC dataset. Table~\ref{tab:ood} shows that our model achieves \textbf{59.45\%} R@1 and \textbf{45.25\%} mAP, significantly outperforming baselines trained on the same data (e.g., RaSa at 54.12\% R@1). This demonstrates that our multi-agent workflow allows the model to learn generalized representations of anomaly intents rather than overfitting to specific dataset biases.

\subsection{Ablation Studies}
\label{sec:ablation}
We conduct a comprehensive ablation study to validate the effectiveness of each module within the Detective Squad and the necessity of our fine-tuning strategy (see Table~\ref{tab:squad_ablation}). The baseline Structure-Aware Coarse Retriever achieves an R@1 of 84.28\%. Regarding the contribution of individual agents, introducing the \textbf{Analyst}  yields a substantial performance gain ($\uparrow$ 1.92\% R@1), significantly outperforming the \textbf{Writer}-only configuration ($\uparrow$ 0.40\%). This empirical evidence underscores that the core bottleneck in bridging the Pose-Semantic Gap is the lack of \textit{explicit visual evidence extraction}. While the Writer can smooth the textual query, it struggles to generate new semantic information without the Analyst's fine-grained observations. Furthermore, the full collaborative squad  achieves the peak performance of \textbf{87.21\%}, demonstrating a strong synergy where the combined agents outperform their individual contributions. Finally, comparing the full squad with LoRA against the zero-shot prompting baseline, we observe a clear performance drop of 1.42\% (87.21\% $\to$ 85.79\%) when LoRA is removed. This indicates that while modern MLLMs possess general knowledge, LoRA fine-tuning is indispensable for aligning the model's attention with specific anomaly definitions and ensuring strict adherence to the multi-agent workflow.

\begin{table}[t]
\centering

\resizebox{\linewidth}{!}{%
\begin{tabular}{c|cccc|cc}
\toprule
\rowcolor{gray!20} 
\textbf{No.} & \textbf{Detective} & \textbf{Analyst} & \textbf{Writer} & \textbf{LoRA} & \textbf{R@1} & \textbf{mAP} \\ \midrule

1 & \no & \no & \no & \no & 84.28 & 91.28 \\ 

2 & \yes & \no & \no & \yes & 84.28 & 91.28 \\ 

3 & \yes & \yes & \no & \yes & 86.20 & 92.34 \\ 

4 & \yes & \no & \yes & \yes & 84.68 & 91.54 \\ 

5 & \yes & \yes & \yes & \no & 85.79 & 92.12 \\ 

6 & \yes & \yes & \yes & \yes & \textbf{87.21} & \textbf{92.87} \\ \bottomrule

\end{tabular}%
}

\caption{\textbf{Ablation study of the Detective Squad Interaction.} We analyze the contribution of each agent role and the impact of the LoRA fine-tuning strategy.}
\label{tab:squad_ablation} 
\end{table}

\subsection{ Impact of Foundation Model Selection.} 
We justify utilizing {Qwen3-VL-8B}\cite{bai2025qwen3vltechnicalreport} as the unified backbone by benchmarking it against the Qwen2.5 series and the state-of-the-art {InternVL3.5-8B}\cite{wang2025internvl35advancingopensourcemultimodal}. As shown in Table~\ref{tab:ablation_models}, while Qwen2.5-VL \cite{bai2025qwen25vltechnicalreport}provides a decent baseline, it lacks the deep reasoning required for subtle anomalies. {InternVL3.5-8B} proves to be a strong competitor with impressive visual understanding (84.65\% R@1). However, {Qwen3-VL-8B} achieves the best overall performance ({84.73\%} R@1). We attribute this superiority to its balanced proficiency in both \textit{visual chain-of-thought} (crucial for the Analyst) and \textit{complex instruction following} (crucial for the Writer). Consequently, we select Qwen3-VL-8B as the optimal single-model engine to drive our collaborative squad.

\begin{table}[t]
\centering

\resizebox{\linewidth}{!}{%
\begin{tabular}{ccc|cc}
\toprule
\multicolumn{3}{c|}{\textbf{Stage-wise Model Selection}} & \multicolumn{2}{c}{\textbf{Performance}} \\ \cmidrule(lr){1-3} \cmidrule(lr){4-5} 
\textbf{\begin{tabular}[c]{@{}c@{}}Detective\end{tabular}} & 
\textbf{\begin{tabular}[c]{@{}c@{}}Analyst \end{tabular}} & 
\textbf{\begin{tabular}[c]{@{}c@{}}Writer  \end{tabular}} & 
\textbf{R@1} & \textbf{mAP} \\ \midrule


Qwen2.5-VL-7B & Qwen2.5-VL-7B & Qwen2.5-7B & 85.69&  92.03 \\ \midrule

InternVL3.5-8B & InternVL3.5-8B & InternVL3.5-8B & 86.10 & 92.19 \\ \midrule

{Qwen3-VL-8B} & {Qwen3-VL-8B} & {Qwen3-VL-8B} & \textbf{86.20} & \textbf{ 92.34}

\\ \bottomrule
\end{tabular}%
}

\caption{{Ablation study on Foundation Model Selection.} }
\label{tab:ablation_models}
\end{table}

\begin{figure}[t]
  \centering
   \includegraphics[width=\linewidth]{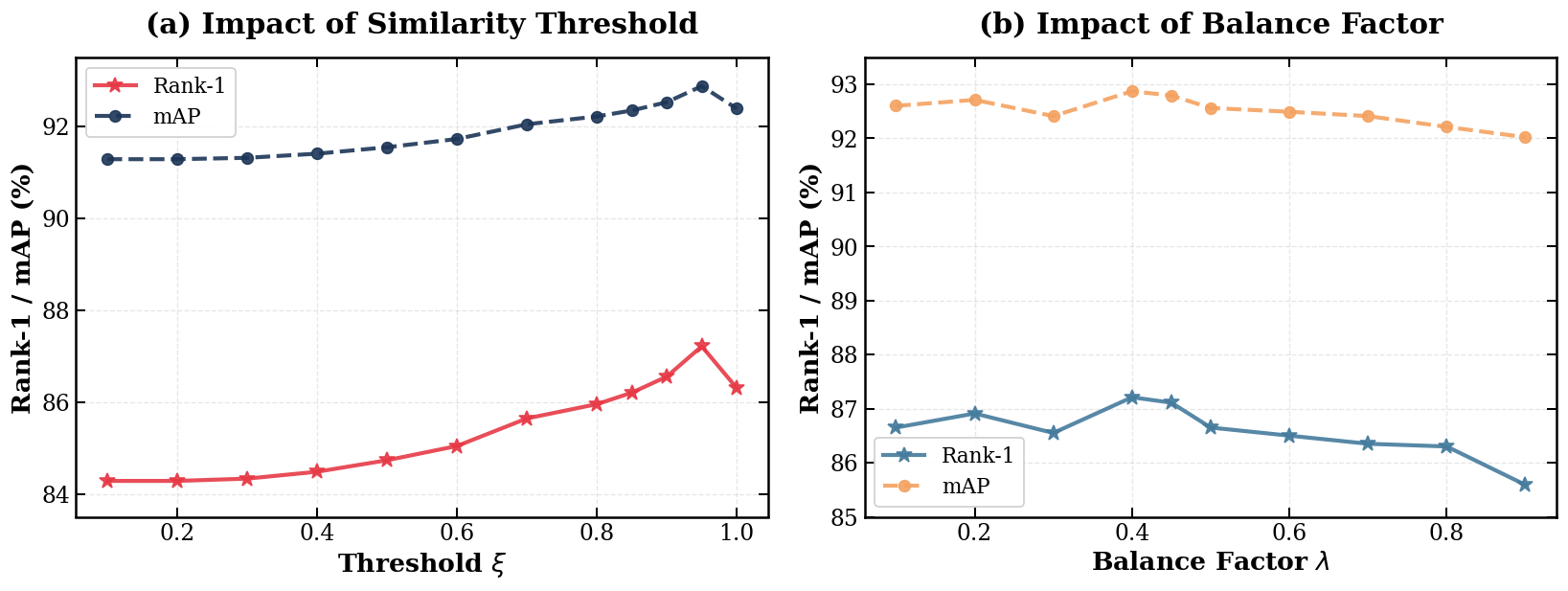}
  \caption{Parameter sensitivity analysis of SSDC.}
  \label{fig:sensitivity}
\end{figure}

\begin{figure}[t]
  \centering
   \includegraphics[width=1.05 \linewidth]{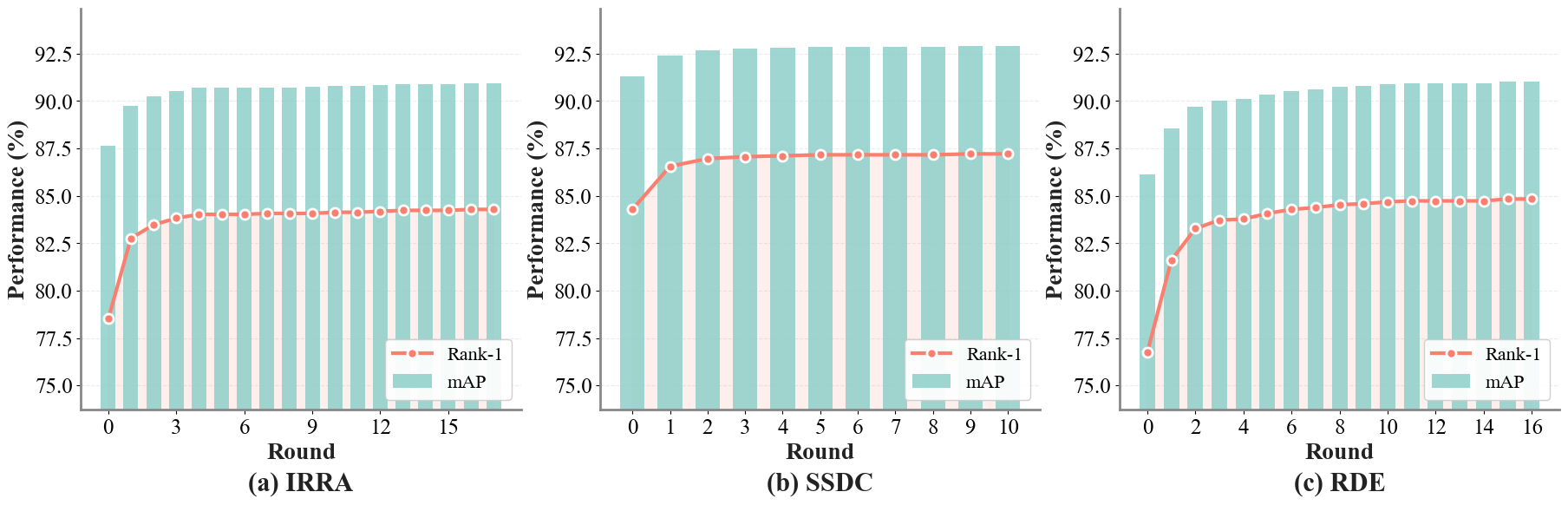}
\caption{{Evolution of Rank-1 and mAP performance versus interaction rounds} for IRRA, SSDC, and RDE. Round 0 denotes the baseline result without the Detective Squad. Subsequent rounds represent the iterative refinement cycles.}
  \label{fig:result}
\end{figure}

\subsection{Efficiency Analysis}
We analyze the trade-off between accuracy and inference cost. Directly applying the multi-agent Detective Squad to the full 1M+ gallery is computationally prohibitive, so we use a cascade strategy for real-time deployment. First, the lightweight CMP model filters out over 99.9\% of candidates and invokes the squad only for top-ranked samples that satisfy a high-confidence threshold ($S_{str} > 0.95$). Second, within the squad, we use \textit{Efficiency-Aware Dynamic Inference}: the {Detective} acts as a gatekeeper, and the computationally intensive {Analyst} and {Writer} run \textit{if and only if} the Detective verifies a ``Match''. This design focuses expensive reasoning on the most ambiguous samples, reducing latency by orders of magnitude while preserving the accuracy of exhaustive re-ranking.

\subsection{Parameter Sensitivity Analysis}

\noindent \textbf{Impact of Similarity Threshold $\xi$.} 
As shown in Figure~\ref{fig:sensitivity}(a), we analyze the threshold $\xi$ which governs the activation of the Detective Squad. We observe a steady improvement in Rank-1 accuracy as the threshold increases, peaking at $\mathbf{\xi=0.95}$. This trend validates our Efficiency-Aware strategy: by setting a high threshold, we effectively filter out structurally irrelevant noise, ensuring that the computationally intensive semantic verification is reserved strictly for ambiguous, high-value candidates that genuinely require fine-grained scrutiny.

\noindent \textbf{Impact of Balance Factor $\lambda$.} 
We further analyze the fusion weight $\lambda$, which balances the structural priors ($S_{str}$) and semantic scores ($S_{sem}$). As illustrated in Figure~\ref{fig:sensitivity}(b), performance maximizes at $\mathbf{\lambda=0.4}$, implying a higher reliance on the semantic score ($1-\lambda=0.6$). This confirms that while structural alignment is essential for initial recall, the detailed reasoning provided by the Detective Squad serves as the decisive factor in resolving the Pose-Semantic Gap. The results remain robust within the range $[0.3, 0.5]$, leading us to set $\lambda=0.4$ as the default.

\subsection{Iterative Reasoning Analysis}

This section further explores the efficacy of the \textbf{Detective Squad} in resolving semantic ambiguities through iterative refinement. We report the performance changes in terms of {mAP} and {Rank-1 (R@1)} in Figure \ref{fig:result}, where \textbf{one interaction round} is defined as a complete inference cycle passing through all three agents .
Experimental results indicate that the most significant improvements in both mAP and R@1 occur in the initial rounds ($\leq 2$). This is attributed to the \textbf{Detective Squad} completing its primary verification loop: the \textbf{Detective} first filters distinct hard negatives, the \textbf{Analyst} extracts fine-grained evidence, and the \textbf{Writer} synthesizes a precise caption to bridge the {Pose-Semantic Gap}. As the number of refinement cycles increases, the performance gains gradually stabilize. This confirms that the initial collaborative pass of the squad effectively resolves the majority of ambiguities, while subsequent rounds provide marginal semantic polishing.




\section{Conclusion}
In this paper, we address the challenge of {Text-based Person Anomaly Search} by leveraging the large-scale {Pedestrian Anomaly Behavior (PAB)} benchmark~\cite{yang2025beyondcmp} to bridge the gap between synthetic training and real-world evaluation. To resolve the inherent ``Pose-Semantic Gap,'' we propose the {Structure-Semantic Decoupled Cascade (SSDC)} framework. By synergizing a lightweight pose-aware retriever with a multi-agent {Detective Squad}, our approach effectively resolves fine-grained ambiguities that traditional methods miss. Extensive experiments confirm that SSDC establishes new state-of-the-art performance, demonstrating remarkable robustness across diverse environmental conditions.


\section*{Acknowledgments}
This work was supported by the National Natural Science Foundation of China under Grant No. U24A20326, and the State Grid Zhejiang Electric Power Cooperation Technology Project (Grant Number: B311DS240012).
\bibliography{main}

\newpage

\newpage

\appendix

\end{document}